\documentclass[11pt]{article}

\usepackage[preprint]{acl}

\usepackage{times}
\usepackage{latexsym}

\usepackage[T1]{fontenc}

\usepackage[utf8]{inputenc}

\usepackage{microtype}

\usepackage{inconsolata}

\usepackage{graphicx}

\usepackage{amsmath,amssymb}
\usepackage{subcaption}
\usepackage{enumitem}
\usepackage{booktabs}
\usepackage{multirow}
\usepackage{xcolor}
\usepackage{makecell}

\author{
  Nathanaël Carraz Rakotonirina\quad Momchil Hardalov \\
  \bfseries Gonzalo Iglesias\quad Adri\`a de Gispert \\
  Amazon AGI \\
  \texttt{\{ncarraz, momchilh, gjii, agispert\}@amazon.com}
}

\title{Where Should a Document Live: \\
Context, Representations, or Parameters?}

\begin{document}
\maketitle
\begin{abstract}
To answer questions outside of their pre-training data, large language models (LLMs) need access to new information, which can be presented in the context window as documents, encoded into the model's \emph{parameters}, or injected as latent \emph{representations}.
However, each of these methods comes with different efficiency, cost, and performance trade-offs, with no single winner.
We present a controlled comparison of representation-based (KV-cache based) and parametric (fine-tuning-based) adaptation methods on five knowledge-intensive benchmarks. We show that in the oracle setting, Cartridges (KV) are the most accurate injection method at nearly every storage budget, outperforming parametric methods by 10 points. Compaction (KV) matches Cartridges only at low compression rates, lagging behind the parametric methods by 10 points at rates higher than $50\times$. In the more realistic multi-document retrieval scenario, Cartridges are the only method that matches in-context learning (ICL), leading the parametric methods by 29 points and Compaction by 15 points. Nonetheless, Cartridges are also the only method, besides full fine-tuning and large MLP adapters, that suffers from catastrophic forgetting, i.e.,~a 6\% performance degradation on control benchmarks, with 13\% in coding.

\end{abstract}

\section{Introduction}
Large language models (LLMs) often need additional information, such as enterprise knowledge bases, patient records, or legal corpora to answer questions that were not part of their training data. Techniques such as in-context learning~\citep[ICL]{NEURIPS2020_1457c0d6} and retrieval-augmented generation~\citep[RAG]{lewis2020retrieval} offer a straightforward solution by placing the document in the context window. However, they are bounded by the context length and require re-processing the entire document for every query.

Knowledge injection methods, on the other hand, encode the documents once into compact 
adapters, replacing the textual context during inference~\citep{eyuboglu2025cartridges, su2025parametric, caccia2025training}. These methods fall into two families: \emph{Representation-based} methods, which incorporate the document into a compressed key--value (KV) cache prefix (e.g. Cartridges~\citep{eyuboglu2025cartridges} and Compaction~\citep{zweiger2026fast}); and \emph{Parametric} methods, which encode the knowledge into the model parameters (e.g. full fine-tuning, LoRA~\citep{hu2022lora}, or MLP adapters~\citep{houlsby2019parameter}). Throughout the paper we use adapters as an umbrella term for both KV caches and model weights.

Even though these methods are widely used in practice, prior work mainly evaluates question-answering (QA) tasks over Wikipedia documents that the model has likely already seen in pre-training, therefore testing recall rather than genuinely encoding new knowledge~\citep{ovadia2024finetuning, su2025parametric}. \citet{eyuboglu2025cartridges} takes a step towards a fairer comparison between parameter-based and representation-based methods, but it covers only a limited set of methods and in a single-document setting. In contrast, in our work we compare head-to-head five state-of-the-art adaptation approaches on five knowledge-intensive tasks, both in the single-document and in the more realistic multi-document retrieval setting. Moreover, we quantify catastrophic forgetting on representative control benchmarks (math, instruction following, general knowledge, coding) and perform a storage-matched and runtime analysis of each method. Our analysis reveals that there is no single best method, and which method to use will depend on the use case. Our main contributions are the following:

\begin{itemize}
    \item We present a holistic, controlled comparison of different knowledge injection methods: (\emph{i})~representation-based (Cartridges, Compaction); and (\emph{ii})~parametric (LoRA, MLP adapters, full fine-tuning) on knowledge-intensive tasks in both single- and multiple-document settings.
    
    \item We study how adapter (KV caches or weights) size  influences model performance and catastrophic forgetting, quantifying the trade-off for each knowledge injection method.
    
    \item We show that adapter composition in a multi-document retrieval setting is possible under certain conditions, but it remains challenging.
\end{itemize}

\section{Related work}

\paragraph{Parameter-Efficient Fine-Tuning.} PEFT methods adapt a model by training only a small subset of its parameters or representations. They can be broadly categorized into two groups. \emph{Representation-based} methods prepend a sequence of trainable vectors to the input and optimize only those vectors, and sometimes their activations across layers \citep{li2021prefix, lester-etal-2021-power, liu2022p, eyuboglu2025cartridges}. \emph{Parametric} methods train a small number of parameters. Representative examples include adapters \citep{houlsby2019parameter}, small trainable modules inserted between layers, and LoRA \citep{hu2022lora}, which learns a low-rank update to the weight matrices. Compared with full fine-tuning, PEFT offers several advantages: it requires less memory and compute to train, it can be stored and served efficiently \citep{chen2023punica}, and multiple modules or adapters can be composed \citep{yadav2023tiesmerging, zhong2024multi, prabhakar2025lora, hardalov2026cartridges}. Finally, it better preserves the model's general capabilities. In particular, LoRA matches the accuracy and sample efficiency of full fine-tuning while being less prone to catastrophic forgetting \citep{schulman2025lora}.

\paragraph{Knowledge injection.} Once Large Language Models (LLMs) are trained, they encode only the knowledge present in their training data. To answer questions about specific documents, a knowledge base, or any other source of knowledge, they must go through an adaptation phase. The most straightforward way is to add the relevant documents to the model's context, possibly preceded by a retrieval stage that selects and filters the documents to be inserted \citep{lewis2020retrieval, guu2020retrieval, gao2023retrieval}. An alternative to in-context injection is to encode the documents directly in the representations \citep{kujanpaa2024knowledge, rakotonirina2024memoryprompt, eyuboglu2025cartridges} or parameters \citep{xiao2023plug, su2025parametric} of the model. Representation-based knowledge injection is akin to prompt compression methods \citep{mu2023learning, chevalier2023adapting, qin2024dodo, lajewska-etal-2025-understanding} that encode the context into shorter sequences of vectors or soft tokens. Recent work \citep{eyuboglu2025cartridges, su2025parametric, caccia2025training} has demonstrated that the best way to inject new knowledge is to train on synthetic data derived from the corpus rather than on the corpus itself, and to use a distillation objective rather than next-token prediction.

\paragraph{Comparing context, representations, and parameters.} Several lines of work compare context-based and parametric knowledge injection, but reach mixed and narrowly scoped conclusions. \citet{ovadia2024finetuning} report that retrieval-augmented generation (RAG), which relies on in-context learning, consistently beats unsupervised fine-tuning for fact injection. However, they do not cover supervised fine-tuning or parameter-efficient adaptation. Parametric RAG \citep{su2025parametric} encodes each document into a separate LoRA adapter and merges the adapters of the retrieved documents at inference time. It reports higher accuracy than in-context RAG in addition to lower latency. Subsequent work \citep{tang2025understanding} shows that parametric methods do not consistently outperform text-based RAG, and that combining the two performs best. Both these works train and evaluate on QA datasets whose documents are Wikipedia paragraphs that were presumably already seen during pretraining, so the model is tested on recalling facts already encoded in its weights rather than on encoding new ones. In contrast, we evaluate on knowledge-intensive tasks with uncontaminated documents, as confirmed by our \emph{No Context} baseline, which queries the LLM without access to external documents. Closest to our setting, \citet{eyuboglu2025cartridges} compare Cartridges against in-context learning and LoRA, but report only limited results for the multi-document case where adapters are composed. We instead run an extensive, controlled comparison that spans all adaptation families and knowledge-intensive datasets varying widely in document length and task type, covering both the single-document setting (only the gold document is encoded) and the more realistic multi-document setting (several documents are retrieved and composed).

\section{Methodology}

\paragraph{Adaptation methods.} We conduct a holistic evaluation on a variety of state-of-the-art adaptation methods from different families: context-based, representation-based, and parametric. We evaluate the following methods:

\begin{itemize}
    \item \textbf{No context}: only the question is given to the model. This baseline quantifies how much can be answered from pretraining knowledge alone, with no access to external documents.
    \item \textbf{ICL}: the document is placed in the model's context. This is the standard in-context baseline. In the multiple document setting, this is standard text-based RAG. 
    \item \textbf{Cartridges} \citep{eyuboglu2025cartridges}: a representation-based method that trains a small KV-cache prefix that is prepended to the question during inference. For the multi-document setting, Cartridges are trained following \citet{hardalov2026cartridges}.
    \item \textbf{Compaction} \citep{zweiger2026fast}: a representation-based method that compresses the document into a compact set of key--value pairs by matching the attention the model would place on the full document. 
    \item \textbf{LoRA} \citep{hu2022lora}: a parametric method that adds trainable low-rank updates to the feed-forward projections of every layer, keeping the base model frozen. 
    \item \textbf{MLP adapters} \citep{houlsby2019parameter}: a parametric method that inserts a small MLP after the attention block of every layer, training only these modules while keeping the base model frozen. 
    \item \textbf{Full fine-tuning}: a parametric method that updates all model parameters, serving as an upper bound on capacity.
\end{itemize}

\paragraph{Knowledge injection settings.} The task we are focusing on is \emph{what is the best approach for injecting new knowledge into a model}. Importantly, we do not limit the QA task to a single-document scenario, where only the gold document is passed along with the questions in the model context, but also consider the more realistic scenario where the model answers based on documents retrieved from the corpus.
In our experiments, we learn one adapter per document and pass only the question and the system instruction into the model prompt. These two settings can be formalized as follows:
\begin{itemize}
    \item \textbf{Single (gold) document}: 
    This is the oracle setting, i.e.,~we encode only the gold document in the model prompt; the goal is to isolate how knowledge in a single document is injected, independent of retrieval.
    \item \textbf{Multiple documents}: This follows the RAG setting, i.e.,~for each question we retrieve the top-$k$ chunks and compose their corresponding adapters. For the representation-based methods we concatenate the $k$ retrieved KV caches, and for the parametric methods we merge the $k$ retrieved weights into a single set of weights. We have compared different merging approaches in Appendix~\ref{sec:merging}, based on which we selected the simple model merge by averaging. We also jointly train on all documents for parametric methods. 
\end{itemize}

\section{Experimental Settings}

\begin{table*}[t!]
\centering
\resizebox{\textwidth}{!}{%
\small
\setlength{\tabcolsep}{3pt} %
\begin{tabular}{lrrrrrll}
\toprule
\textbf{Dataset} & \textbf{Docs} & \textbf{Questions} & \textbf{Qs/Doc} & \textbf{Avg.\ Tok.} & \textbf{Total Tok.} & \textbf{Task Type} & \textbf{Domain} \\
\midrule
LongHealth   & 20    & 400   & 20.0 & 11,700 &236K &  Multiple-choice (5-way) & Clinical patient records \\
QASPER       & 407   & 1,451 & 3.5 & 4,751  & 	665K &  Extract./free-form/yes-no & Full research papers \\
QuALITY      & 115   & 2,086 & 18.1 & 5,713  & 1.9M & Multiple-choice (4-way) & Fiction \& non-fiction narratives \\
T$^2$-RB/FinQA & 380 & 1,147 & 3.0 & 1,026 & 	392K & Math. Calculation & Corporate earnings reports w/ tables \\
TechQA       & 496   & 610   & 1.8 & 1,509  & 748K& Extractive & IBM IT support technotes \\
\bottomrule
\end{tabular}%
}
\caption{Dataset statistics where \textbf{Docs} is the number of unique documents (one cartridge per document), \textbf{Qs/Doc} is the average number of questions per document, and \textbf{Avg.\ Tok.} is the average document length in tokens (Qwen3-8B tokenizer).}
\label{tab:datasets}
\end{table*}

\paragraph{Models.} We use Qwen3-8B \citep{yang2025qwen3} as our main model. We also validate our results on Gemma-3-12B \citep{gemma3_report} in Appendix \ref{sec:gemma}.

\paragraph{Dataset.} We evaluate on the following datasets (see Table \ref{tab:datasets} for dataset statistics) using their corresponding metric:\footnote{We adopted the datasets and metrics as described in \citet{hardalov2026cartridges}.}

\begin{itemize}[nosep]
    \item \textbf{LongHealth} \cite{adams2024longhealth}: A clinical QA benchmark consisting of detailed fictional patient records with multiple-choice questions that test information extraction, negation understanding, and temporal sorting over long medical notes. We extract the model's answer by fuzzy option matching, then compute exact match against the gold option.
    \item \textbf{QASPER} \cite{dasigi2021qasper}: An info-seeking QA dataset over full NLP research papers, where questions are written by NLP practitioners who saw only the title and abstract. Answers include extractive spans, free-form text, and yes/no responses. The metric is token-level F1 with the reference answers.
    \item \textbf{QuALITY} \cite{pang2022quality}: A multiple-choice reading comprehension benchmark over long-form fiction and non-fiction narratives, where questions require careful reading and reasoning rather than surface-level pattern matching. The metric is exact match on the extracted answer letter (A, B, C, or D).
    \item \textbf{T$^2$-RAGBench/FinQA} \cite{strich2026t2ragbench,chen2021finqa}: A decontextualized variant of FinQA designed for RAG evaluation, where questions about corporate earnings reports have been rewritten to be context-independent. The task requires numerical reasoning by constructing mathematical formulas over tables and text extracted from financial filings. The model is expected to produce a formula, which we evaluate to a numerical value and compare against the ground-truth answer with a relative tolerance of 1\%.
    \item \textbf{TechQA} \cite{castelli2020techqa}: A technical support QA dataset drawn from IBM's Technote corpus, where questions require extracting precise solutions from IT documentation covering enterprise software and infrastructure issues. We use DeepSeek-Distilled-Qwen-32B \cite{guo2025deepseek} as a judge to evaluate the model's prediction against the reference answer (see Appendix \ref{sec:judge} for more details).
\end{itemize}

To assess whether knowledge injection degrades the model's general abilities, we additionally evaluate on four control benchmarks: GSM8K \cite{cobbe2021gsm8k} for grade-school math, HumanEval \cite{chen2021humaneval} for code generation, IFEval \cite{zhou2023ifeval} for instruction following, and MMLU \cite{hendrycks2021mmlu} for broad knowledge. We evaluate adapted models and compare against the original model to measure catastrophic forgetting (Section~\ref{analysis}).

\paragraph{Training data.} Following \citet{eyuboglu2025cartridges} and \citet{hardalov2026cartridges}, we train on LLM-generated synthetic data called \textit{Self-Study} data. More concretely, a question-generator LLM receives a part of the original document and a seed prompt, and produces a set of questions. An answer-generator LLM, given the same  document, but not the seed prompt, answers each question providing tokens and their log probability distributions. Same as \citet{hardalov2026cartridges} we use a GPT-OSS 120B~\citep{agarwal2025gpt} as the question generator and our target model as the answer generator. We generate $n=20$ questions per chunk (more details in Appendix~\ref{sec:synthesis_details}). We train all methods using the same dataset.

\begin{table*}[t!]
\centering
{
\begin{tabular}{lcccccc}
\toprule
\textbf{Method} & \textbf{LongHealth} & \textbf{QuALITY} & \textbf{QASPER} & \textbf{FinQA} & \textbf{TechQA} & \textbf{Avg.} \\
 & (Acc.) & (Acc.) & (F1) & (EM) & (Judge) & \\
\midrule
No context         & 37.5 {\scriptsize $\pm$ 1.1} & 43.6 {\scriptsize $\pm$ 0.4} & 19.2 {\scriptsize $\pm$ 0.6} & 2.9 {\scriptsize $\pm$ 0.0} & 21.1 {\scriptsize $\pm$ 1.1} & 24.9 \\
ICL       & \underline{87.4} {\scriptsize $\pm$ 0.8} & \textbf{82.5} {\scriptsize $\pm$ 0.3} & \textbf{56.7} {\scriptsize $\pm$ 0.4} & \textbf{66.8} {\scriptsize $\pm$ 2.7} & 74.7 {\scriptsize $\pm$ 0.9} & \textbf{73.6} \\
\cmidrule(l){1-7}
Cartridge    & 81.1 {\scriptsize $\pm$ 1.1} & 78.6 {\scriptsize $\pm$ 0.9} & \underline{54.9} {\scriptsize $\pm$ 0.3} & 62.7 {\scriptsize $\pm$ 0.2} & \underline{75.8} {\scriptsize $\pm$ 0.7} & 70.6 \\
Compaction    & \textbf{87.7} {\scriptsize $\pm$ 0.9} & \underline{82.1} {\scriptsize $\pm$ 0.3} & 54.8 {\scriptsize $\pm$ 0.1} & \underline{66.4} {\scriptsize $\pm$ 0.2} & \textbf{76.0} {\scriptsize $\pm$ 2.4} & \underline{73.4} \\
\cmidrule(l){1-7}
LoRA               & 75.3 {\scriptsize $\pm$ 0.9} & 73.7 {\scriptsize $\pm$ 1.2} & 50.3 {\scriptsize $\pm$ 0.6} & 49.0 {\scriptsize $\pm$ 0.9} & 74.3 {\scriptsize $\pm$ 2.6} & 64.5 \\
MLP adapters       & 74.8 {\scriptsize $\pm$ 0.9} & 72.3 {\scriptsize $\pm$ 1.2} & 47.5 {\scriptsize $\pm$ 0.9} & 44.1 {\scriptsize $\pm$ 0.3} & 72.5 {\scriptsize $\pm$ 1.8} & 62.2 \\
Full fine-tuning   & 69.0 {\scriptsize $\pm$ 2.4} & 72.8 {\scriptsize $\pm$ 0.3} & 39.8 {\scriptsize $\pm$ 0.3} & 41.6 {\scriptsize $\pm$ 2.8} & \textbf{76.0} {\scriptsize $\pm$ 3.4} & 59.8 \\
\bottomrule
\end{tabular}%
}
\caption{\textbf{Single-document scores at a fixed adapter size (Qwen3-8B).} Cartridge and Compaction use $2\times$ compression, LoRA rank $64$, and MLP a bottleneck of $512$. Scores are averaged across 3 runs ($\pm$ standard deviation); \textbf{bold} marks the best method per dataset and \underline{underline} the second best.}
\label{tab:single-doc}
\end{table*}

\paragraph{Objective function.} We use the distillation objective that minimizes the KL divergence between a teacher, which is a model with the document in context, and a student, which is the same model to be adapted. We select this distillation objective since it achieves a better performance compared to next-token-prediction for knowledge injection \citep{eyuboglu2025cartridges}. We corroborate these findings for LoRA in Appendix~\ref{sec:distill-ntp}. 

\paragraph{Retrieval.}
For the multi-document setting, we index the documents into chunks and retrieve the relevant chunks for each question, without query reformulation. For ICL and representation-based methods, the documents or KV caches are concatenated following their retrieval order. With 1,024-token chunks, LongHealth and QuALITY average ${\sim}2$ chunks per unique adapter, while for TechQA and FinQA the ratio is closer to 1:1 (see Table~\ref{tab:unique_docs}). We provide more details about retrieval and the RAG baseline in Appendix \ref{sec:rag_details}.

\paragraph{Hyperparameters.} For the fixed adapter size experiments, we use a compression rate of $2\times$ for Cartridge and Compaction, a rank of $64$ (with $\alpha=128$) for LoRA, and a bottleneck dimension of $512$ for MLP adapters. LoRA is applied to the feed-forward projections of every layer, and the MLP adapters are added as a residual bottleneck on the output of every transformer layer. When comparing the methods across a range of adapter sizes, we vary the compression rate over $2\times$, $10\times$, $20\times$, $50\times$, and $100\times$. We then choose the LoRA rank and the MLP bottleneck dimension so that each adapter matches the per-document memory footprint of the KV-cache methods at that rate. Because storage is tied to document length, these matched sizes are dataset-dependent. We do not report a storage-matched adapter at $100\times$ because the target footprint falls below the smallest trainable adapter (a rank of at least one). During inference, we use the decoding hyperparameters recommended for Qwen3 \citep{yang2025qwen3}: temperature $0.6$, top-$p$ $0.95$, and top-$k$ $20$, with a maximum of $4{,}096$ generated tokens. We list full training hyperparameters in Appendix~\ref{sec:hyperparameters}.

\section{Experimental Results}
\label{sec:results}

\begin{figure*}[t!]
\centering
\includegraphics[width=\textwidth]{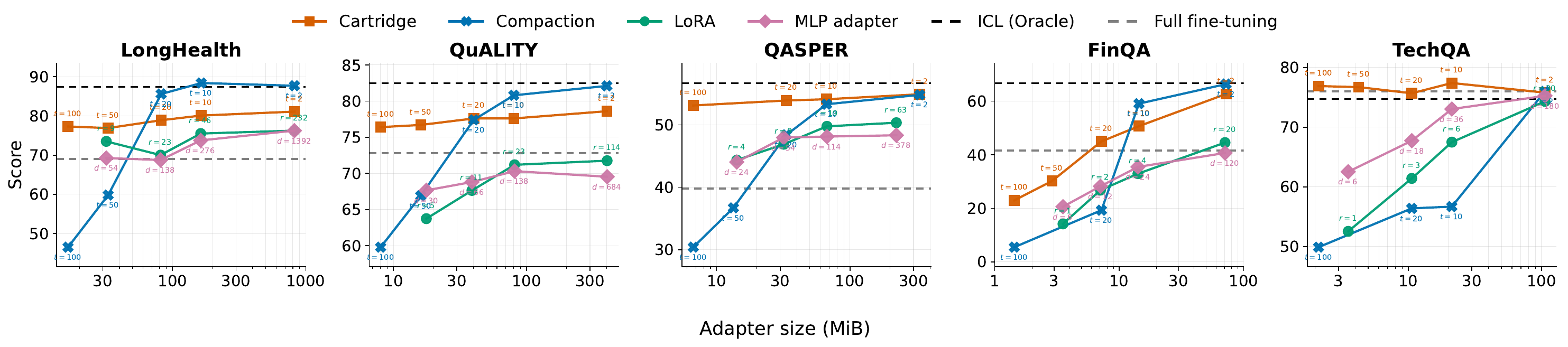}
\caption{\textbf{Single-document score versus adapter size} (MiB, log scale). Cartridges and Compaction are swept over compression rates $2\times$, $10\times$, $20\times$, $50\times$, and $100\times$. LoRA and the MLP adapter use the rank and bottleneck dimension that match the KV-cache memory footprint at each rate.}
\label{fig:storage-pareto}
\end{figure*}

\subsection{Single Oracle document}
\paragraph{Fixed-size adapter.}
Table~\ref{tab:single-doc} shows the model performance in oracle setting, i.e.,~providing only the encoded gold document. \emph{No Context} (24.9) performs far below the other methods, confirming that the knowledge is genuinely injected rather than recalled. This gap is largest on FinQA, where answering a question requires extracting specific numbers from the text and tables of the document. In this setting, representation-based methods have a clear edge and outperform the parametric ones on nearly all datasets. \emph{Compaction} is strongest overall with an average score of 73.4, matching the ICL upper bound of 73.6, followed by Cartridges (70.6). On the parametric side, \emph{LoRA} is the best method (64.5 on average), ahead of \emph{MLP adapters} (62.2) and \emph{full fine-tuning} (59.8). Full fine-tuning is only competitive on TechQA and weakest on datasets whose metric penalizes malformed output (QASPER F1, FinQA formulas), which we relate to catastrophic forgetting (studied in \S~\emph{Catastrophic forgetting}).

\paragraph{Comparing adapter sizes.}
We also compare the methods across a range of adapter sizes in the single-document setting in Figure~\ref{fig:storage-pareto}. The two representation-based methods behave very differently as the compression rate increases. Compaction performs best at low compression but degrades steeply, overtaken by the parametric methods at high compression: on FinQA it falls from $66.4$ at $2\times$ to $19.3$ at $20\times$, and on LongHealth from $87.7$ to $46.5$ at $100\times$. Cartridges instead stay nearly flat (e.g.\ LongHealth $81.1\rightarrow77.3$, QuALITY $78.6\rightarrow76.4$, TechQA $75.8\rightarrow76.9$ from $2\times$ to $100\times$) and dominate the storage-matched parametric methods on every dataset. LoRA and MLP adapter improve with size up to a point and then plateau. We confirm this by training wider adapters in Appendix~\ref{sec:hyperparameters}.

\begin{figure*}[t!]
\centering
\includegraphics[width=\textwidth]{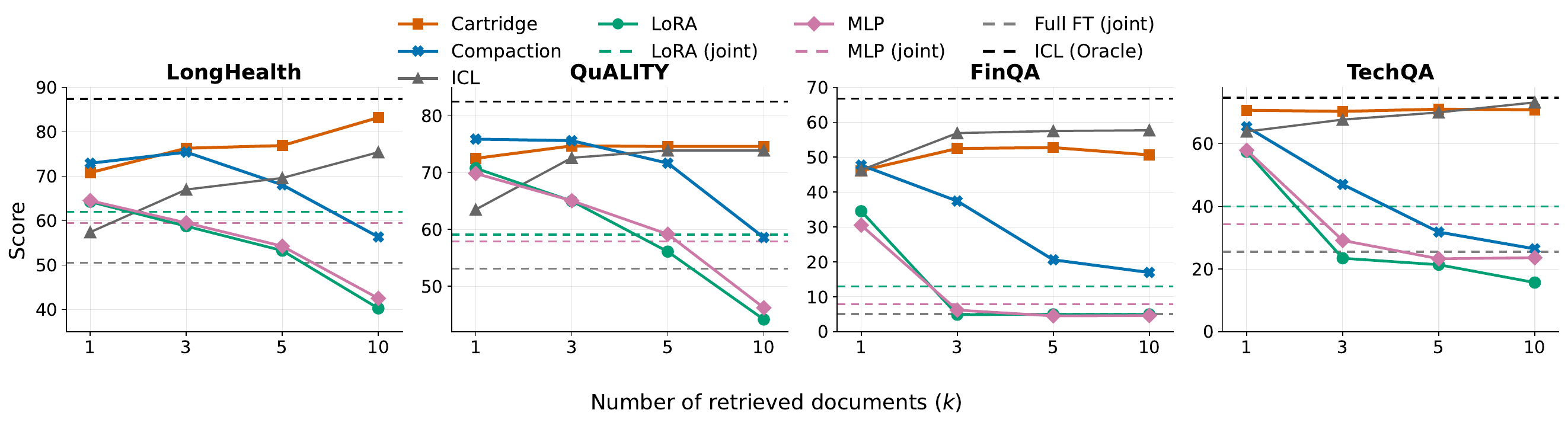}

\includegraphics[width=\textwidth]{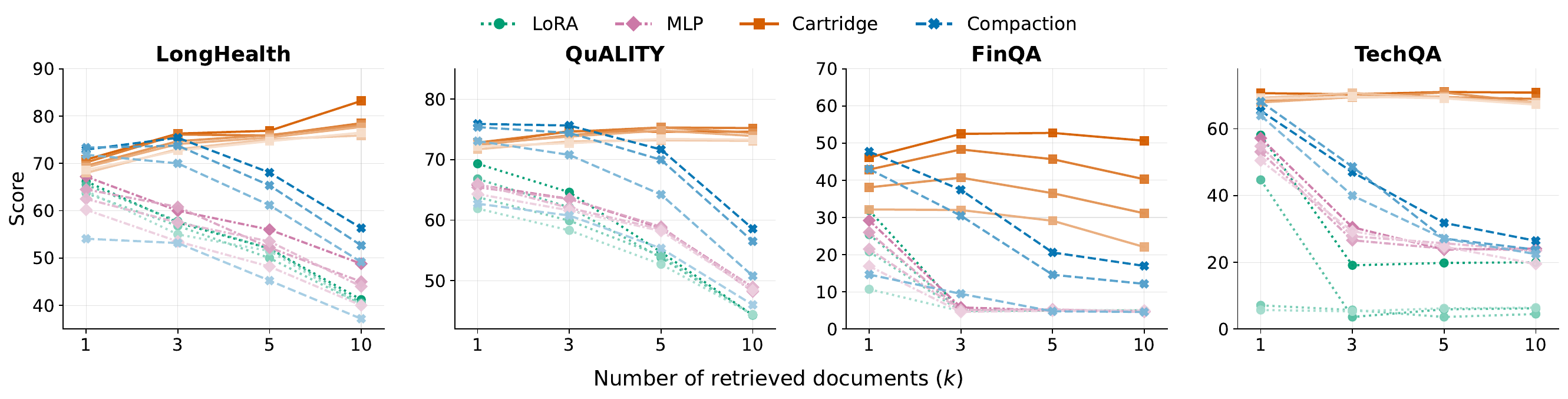}
\caption{\textbf{Multi-document results (Qwen3-8B).} Score versus the number of retrieved documents $k$, composing the top-$k$ retrieved per-document artifacts. \textbf{(Top)} \textbf{Fixed adapter size.} \emph{ICL} refers to RAG in this setting, \emph{ICL (Oracle)} is the in-context upper bound with the gold document in context, and \emph{Joint} is a single adapter trained on all documents. \textbf{(Bottom) Varying adapter size}. Each method is swept over the storage-matched compression ratios (where available), with method encoded by color and compression rate by shade (dark = larger adapter / lower compression).}
\label{fig:multidoc}
\label{fig:rag_vs_k}
\label{fig:lora-storage-topk}
\end{figure*}

\subsection{Multiple retrieved documents} 
\paragraph{Fixed-size adapter.} We retrieve the top-$k$ chunks for each question and compose their adapters (mapping each chunk to its source document and removing duplicates): KV caches are concatenated for the representation-based methods, and weights are averaged for the parametric methods. Figure~\ref{fig:rag_vs_k} reports performance against the number of retrieved documents ($k=1,3,5,10$). In the comparison, we include a single adapter, trained jointly on all documents only for the parametric methods, as \citet{hardalov2026cartridges} shows that Cartridges perform worse when trained jointly, while Compaction shifts the rope embedding to the original size of the compacted caches, which goes beyond the context window of the model. ICL and Cartridges maintain or improve their performance as $k$ increases: from $k{=}1$ to $k{=}10$, Cartridges rise on LongHealth ($70.8\rightarrow83.2$) and hold on TechQA ($70.7\rightarrow70.9$) and QuALITY ($72.5\rightarrow74.6$). In contrast, both Compaction and the merged parametric adapters degrade monotonically. Compaction drops from $k{=}1$ to $k{=}10$ on every dataset (LongHealth $72.9\rightarrow56.3$, TechQA $65.4\rightarrow26.4$), and the merged adapters fall even faster: LoRA on TechQA collapses from $57.4$ at $k{=}1$ to $23.5$ at $k{=}3$, and on QuALITY from $70.8$ to $44.2$ at $k{=}10$. The degradation is most severe on FinQA, where merging collapses accuracy from $34.5$ at $k{=}1$ to $4.9$ at $k{=}3$. Joint training outperforms merging as soon as a single distractor is added (e.g.\ FinQA $13.0$ vs.\ $4.9$ and TechQA $40.0$ vs.\ $23.5$ at $k{=}3$), except on QuALITY, where it takes the lead only at $k{=}5$. Compaction and parametric injection are thus effective in the single-document setting but do not support multi-document composition.

\begin{figure*}[t!]
\centering
\includegraphics[width=\textwidth]{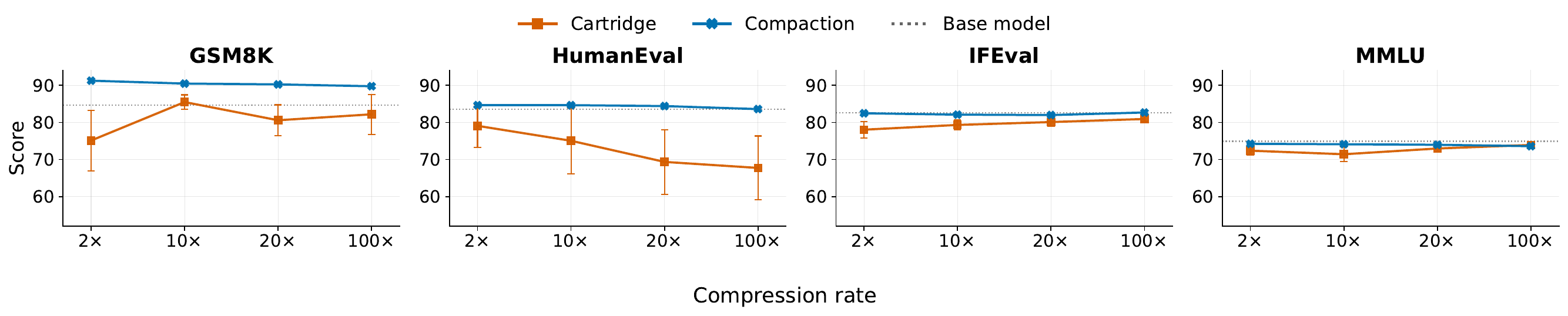}

\includegraphics[width=\textwidth]{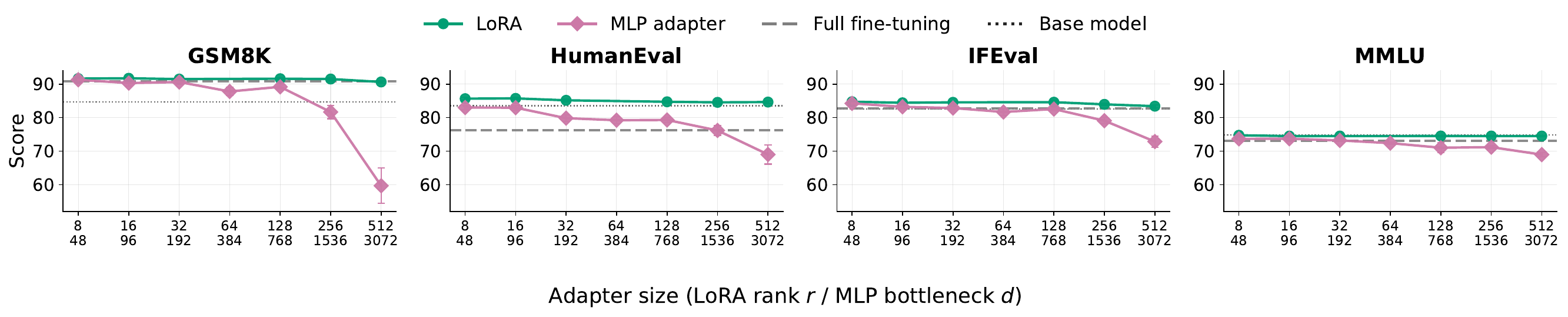}
\caption{\textbf{Forgetting on control benchmarks (Qwen3-8B).} The score is averaged across the five source datasets, and error bars are the standard error across those datasets. \textbf{(Top)} The representation-based methods are swept over compression rate; the lowest-compression column merges the $2\times$ and $3\times$ points (the two longest datasets use $3\times$). \textbf{(Bottom)} The parametric methods are swept over adapter size.}
\label{fig:forgetting}
\end{figure*}

\paragraph{Varying adapter size.} We repeat the multi-document analysis across adapter sizes in Figure~\ref{fig:lora-storage-topk}. Cartridges are mostly stable across compression rates; on LongHealth, for example, the $k{=}10$ score is $83.2$ at $2\times$ and still $76.5$ at $100\times$. The only exception is FinQA, where heavier compression degrades performance ($50.7$ down to $22.1$ from $2\times$ to $20\times$), as in the single-document case. For LoRA and MLP, adapter size has essentially no effect on the composed result: by $k{=}10$ their scores converge regardless of the rate, to around $5$ points on FinQA and $40$--$49$ on LongHealth at both $2\times$ and $50\times$. Compaction behaves similarly under composition but is further hurt by compression, dropping on LongHealth at $k{=}10$ from $56.3$ at $2\times$ to $37.2$ at $50\times$. In short, using a larger rank, a wider bottleneck, or a lower compression rate does not fix the composition problem.

\section{Analysis}
\label{analysis}

\paragraph{Catastrophic forgetting.}
We assess how knowledge injection affects the general capabilities of the model by evaluating the adapted models on four control benchmarks, sampling 3 adapters (an adapter encodes a single document) per training dataset. We plot performance as a function of compression rate for the representation-based methods (Figure~\ref{fig:forgetting}, top) and as a function of adapter size for the parametric methods (Figure~\ref{fig:forgetting}, bottom). Among the representation-based methods, Compaction stays at the base model at every compression rate, while Cartridges degrade by 6\% on average, mainly on code generation, where HumanEval drops by 16 points at high compression. In the parametric methods, LoRA retains the base model's capabilities at every rank, whereas the MLP adapter is stable up to a moderate bottleneck ($d{\le}384$) and then degrades as it widens: on GSM8K it falls from ${\sim}92$ at the smallest bottleneck to ${\sim}60$ at the largest, with parallel drops on every benchmark. The large MLP adapters are trained at a lower learning rate ($2\times10^{-5}$ instead of $10^{-4}$), which we found necessary to avoid a much steeper collapse (e.g.,~GSM8K down to $24$ at $10^{-4}$). LoRA is more robust at matched parameter count, pointing to the low-rank constraint, rather than the number of parameters alone, as what accounts for preserving general capability. 

\paragraph{Cost analysis.}
At inference time, ICL processes all the retrieved documents for each query during the prefill phase\footnote{Prefilling cost can be attenuated over time with prompt caching, which is also applicable to the representation-based methods.} and then decodes while attending over all of them. This can be a large recurring cost, from about $1$k tokens on FinQA to $11$k on LongHealth for the document alone, paid on every query. Representation-based and parametric knowledge injection methods remove the document from the prompt, prefilling only the question. The representation-based methods still attend over a stored prefix, but a compressed one, so their per-query cost shrinks with the compression rate. With representation-based methods a $10\times$ token reduction yields $\sim100\times$ fewer prefill FLOPs due to quadratic attention. At inference, adapters are often pre-computed BF16 tensors, so loading requires only a GPU memory transfer, not a forward pass. On Qwen3-8B with H200 GPUs, loading 10 cartridges at $20\times$ compression (6K KV tokens, 850 MiB) takes 30–50 ms, compared with 400–800 ms to prefill the equivalent 120K raw tokens. For the parametric methods the model attends to only the question, as the document lives in the weights. Overall, every injection method is cheaper per query than ICL, with the parametric methods the cheapest at inference and independent of document length.

\section{Discussion}

\paragraph{Which method to use.} We find that no single method dominates across all considered settings. Cartridges are the most accurate knowledge injection method, including in the more realistic multi-document setting, but they degrade the model's general capabilities at high compression rates. Compaction does not suffer from any forgetting but only matches Cartridges in the single-document setting and at low compression. The parametric methods lag behind Cartridges irrespective of the adapter size. Their advantage is instead that the adapter size does not depend on the document and that they are less prone to forgetting, except for large MLP adapters, which are never chosen in practice as they do not bring any performance boost. Jointly fine-tuning on all documents outperforms merging the parametric adapters when the number of retrieved documents is high. Full fine-tuning is dominated throughout: it is the most expensive to train and store yet the least accurate, and it forgets more than any low-rank adapter.

\paragraph{Composition is not straightforward.} Combining per-document artifacts does not always work out of the box. This holds across families except for Cartridges: merging low-rank or full-rank adapters collapses with even a single distractor, and concatenating several independently compressed caches degrades in the same way. Although \citet{su2025parametric} showed promising results with merging LoRA adapters as an alternative to RAG, their findings were limited to small Wikipedia snippets, and the approach does not hold for the more complex tasks we consider.
The composition of Cartridges in \citep{eyuboglu2025cartridges} was also limited, but the mixed training proposed in \citet{hardalov2026cartridges} allows cartridges to be concatenated and perform well in the multi-document setting. A class of methods called \textit{KV-cache reuse} addresses the problem of concatenating independently compressed caches, and could be applied to Compaction. \emph{KV Packet}~\citep{chen2026kv} and \emph{KVLink}~\citep{yang2026kvlink} add trainable soft tokens to bridge the discontinuities between the caches. \emph{C$^2$KV}~\citep{du2026c} instead compresses and composes jointly, learning a compression objective under which the caches remain mergeable rather than optimizing each cache alone. For the parametric adapters, we are not aware of any competitive approach to combine per-document weights: more advanced weight-merging techniques do not lead to any significant improvement (Appendix~\ref{sec:merging}). Joint fine-tuning is the best available option but still lags composed cartridges, especially on information-dense datasets like FinQA.

\paragraph{On catastrophic forgetting.} 
Among the representation-based methods, only Cartridges suffer from catastrophic forgetting on HumanEval. This suggests that the forgetting does not inherently come from the compressed KV caches. A possible fix is to initialize the KV caches using Compaction during Cartridges training. Regarding the parametric methods, only the full-rank MLP adapter forgets despite using the same number of parameters for all methods. Full fine-tuning also severely degrades the model's general capabilities, suggesting that it is the full-rank nature of the update, rather than the parameter budget, that drives catastrophic forgetting. The low-rank constraint thus acts as a regularizer, suggesting that any parametric injection method intended to preserve general ability should be built around a subspace constraint.

\section{Conclusion}
We present a controlled comparison of representation-based and parametric knowledge injection methods across five knowledge-intensive benchmarks, in both single- and multi-document settings. We find that no method wins on every axis, but the trade-offs are clear. Cartridges are the most accurate injection method at nearly every storage budget and the only one that extends to multiple retrieved documents, at the cost of mild forgetting on the control benchmarks. Compaction matches the in-context oracle at low compression but degrades quickly as compression increases. The parametric adapters trail Cartridges at matched storage and cannot be composed across documents, but they are fixed-size and cheap to serve. Composing independently trained per-document artifacts remains the main open problem; for the representation-based family, compressed-cache reuse is a promising route.

\section*{Limitations} Our study focuses on two mid-sized instruction-tuned models, Qwen3-8B and Gemma-3-12B, and we do not test whether the same trade-offs hold for larger models. We also only investigate knowledge-intensive tasks and do not cover acquiring or improving skills, such as reasoning or coding, which may interact differently with each injection method. Our multi-document analysis relies on a single retrieve-then-compose pipeline: mean-merging for the parametric adapters, and prompt concatenation for the representation-based methods. The composition findings should therefore be read as the behavior of these standard merging operators, which more sophisticated reranking, filtering \citep{asai2024self}, routing, retrieval, or merging schemes may improve. All methods are trained on the same self-study synthetic data, so their relative performance is tied to the quality of that data and of the generators that produce it.

\section*{Ethical Considerations}
Our work is a controlled comparison rather than a deployed system, but the knowledge injection methods we study carry the usual risks of language-model applications. Because the injected knowledge is stored in the weights or KV caches, it is harder to inspect or update than an explicitly retrieved passage, and an adapted model can still hallucinate. All datasets we use are publicly available and, to the best of our knowledge, free of personal data, and we use each of them in accordance with its license (see Appendix~\ref{appx:licenses}). Our training data is generated by an LLM (self-study) and is not additionally filtered. Finally, the catastrophic forgetting we measure is itself a safety-relevant failure mode, as injecting new knowledge can silently erode the general capabilities of the model.

\bibliography{custom}

@inproceedings{eyuboglu2025cartridges,
  title        = {Cartridges: Lightweight and general-purpose long context representations via self-study},
  author       = {Sabri Eyuboglu and Ryan Saul Ehrlich and Simran Arora and Neel Guha and Dylan Zinsley and Emily Ruoyu Liu and Atri Rudra and James Y. Zou and Azalia Mirhoseini and Christopher Re},
  year         = 2025,
  booktitle    = {ES-FoMo III: 3rd Workshop on Efficient Systems for Foundation Models},
  url          = {https://openreview.net/forum?id=DuVSIWY5vC},
}

@article{hardalov2026cartridges,
  title={Cartridges at Scale: Training Modular KV Caches over Large Document Collections},
  author={Hardalov, Momchil and Iglesias, Gonzalo and de Gispert, Adri{\`a}},
  journal={arXiv preprint arXiv:2606.04557},
  year={2026}
}

@inproceedings{liu2022p,
  title={P-tuning: Prompt tuning can be comparable to fine-tuning across scales and tasks},
  author={Liu, Xiao and Ji, Kaixuan and Fu, Yicheng and Tam, Weng and Du, Zhengxiao and Yang, Zhilin and Tang, Jie},
  booktitle={Proceedings of the 60th Annual Meeting of the Association for Computational Linguistics (Volume 2: Short Papers)},
  pages={61--68},
  year={2022}
}

@article{hu2022lora,
  title={Lora: Low-rank adaptation of large language models.},
  author={Hu, Edward J and Shen, Yelong and Wallis, Phillip and Allen-Zhu, Zeyuan and Li, Yuanzhi and Wang, Shean and Wang, Liang and Chen, Weizhu and others},
  journal={Iclr},
  volume={1},
  number={2},
  pages={3},
  year={2022}
}

@article{kujanpaa2024knowledge,
  title={Knowledge injection via prompt distillation},
  author={Kujanp{\"a}{\"a}, Kalle and Valpola, Harri and Ilin, Alexander},
  journal={arXiv e-prints},
  pages={arXiv--2412},
  year={2024}
}

@article{zhong2024multi,
  title={Multi-lora composition for image generation},
  author={Zhong, Ming and Shen, Yelong and Wang, Shuohang and Lu, Yadong and Jiao, Yizhu and Ouyang, Siru and Yu, Donghan and Han, Jiawei and Chen, Weizhu},
  journal={arXiv preprint arXiv:2402.16843},
  year={2024}
}

@inproceedings{prabhakar2025lora,
  title={Lora soups: Merging loras for practical skill composition tasks},
  author={Prabhakar, Akshara and Li, Yuanzhi and Narasimhan, Karthik and Kakade, Sham and Malach, Eran and Jelassi, Samy},
  booktitle={Proceedings of the 31st International Conference on Computational Linguistics: Industry Track},
  pages={644--655},
  year={2025}
}

@article{gemma3_report,
  title={Gemma 3 Technical Report},
  author={Gemma Team},
  journal={arXiv preprint arXiv:2503.19786},
  year={2025},
  url={https://arxiv.org/abs/2503.19786}
}

@inproceedings{xiao2023plug,
  title={Plug-and-play document modules for pre-trained models},
  author={Xiao, Chaojun and Zhang, Zhengyan and Han, Xu and Chan, Chi-Min and Lin, Yankai and Liu, Zhiyuan and Li, Xiangyang and Li, Zhonghua and Cao, Zhao and Sun, Maosong},
  booktitle={Proceedings of the 61st Annual Meeting of the Association for Computational Linguistics (Volume 1: Long Papers)},
  pages={15713--15729},
  year={2023}
}

@article{caccia2025training,
  title={Training plug-n-play knowledge modules with deep context distillation},
  author={Caccia, Lucas and Ansell, Alan and Ponti, Edoardo and Vuli{\'c}, Ivan and Sordoni, Alessandro},
  journal={arXiv preprint arXiv:2503.08727},
  year={2025}
}

@article{mu2023learning,
  title={Learning to compress prompts with gist tokens},
  author={Mu, Jesse and Li, Xiang and Goodman, Noah},
  journal={Advances in Neural Information Processing Systems},
  volume={36},
  pages={19327--19352},
  year={2023}
}

@inproceedings{qin2024dodo,
  title={Dodo: Dynamic contextual compression for decoder-only lms},
  author={Qin, Guanghui and Rosset, Corby and Chau, Ethan and Rao, Nikhil and Van Durme, Benjamin},
  booktitle={Proceedings of the 62nd Annual Meeting of the Association for Computational Linguistics (Volume 1: Long Papers)},
  pages={9961--9975},
  year={2024}
}

@inproceedings{chevalier2023adapting,
  title={Adapting language models to compress contexts},
  author={Chevalier, Alexis and Wettig, Alexander and Ajith, Anirudh and Chen, Danqi},
  booktitle={Proceedings of the 2023 Conference on Empirical Methods in Natural Language Processing},
  pages={3829--3846},
  year={2023}
}

@inproceedings{rakotonirina2024memoryprompt,
  title={MemoryPrompt: A Light Wrapper to Improve Context Tracking in Pre-trained Language Models},
  author={Rakotonirina, Nathana{\"e}l Carraz and Baroni, Marco},
  booktitle={Proceedings of the 2024 Joint International Conference on Computational Linguistics, Language Resources and Evaluation (LREC-COLING 2024)},
  pages={11187--11195},
  year={2024}
}

@inproceedings{su2025parametric,
  title={Parametric retrieval augmented generation},
  author={Su, Weihang and Tang, Yichen and Ai, Qingyao and Yan, Junxi and Wang, Changyue and Wang, Hongning and Ye, Ziyi and Zhou, Yujia and Liu, Yiqun},
  booktitle={Proceedings of the 48th International ACM SIGIR Conference on Research and Development in Information Retrieval},
  pages={1240--1250},
  year={2025}
}

@article{chen2023punica,
  title={Punica: multi-tenant lora serving. ArXiv},
  author={Chen, L and Ye, Z and Wu, Y and Zhuo, D and Ceze, L and Krishnamurthy, A},
  journal={arXiv preprint arXiv:2310.18547},
  year={2023}
}

@article{schulman2025lora,
  author = {John Schulman and Thinking Machines Lab},
  title = {LoRA Without Regret},
  journal = {Thinking Machines Lab: Connectionism},
  year = {2025},
  note = {https://thinkingmachines.ai/blog/lora/},
  doi = {10.64434/tml.20250929},
}

@inproceedings{houlsby2019parameter,
  title={Parameter-efficient transfer learning for NLP},
  author={Houlsby, Neil and Giurgiu, Andrei and Jastrzebski, Stanislaw and Morrone, Bruna and De Laroussilhe, Quentin and Gesmundo, Andrea and Attariyan, Mona and Gelly, Sylvain},
  booktitle={International conference on machine learning},
  pages={2790--2799},
  year={2019},
  organization={PMLR}
}

@inproceedings{lester-etal-2021-power,
    title = "The Power of Scale for Parameter-Efficient Prompt Tuning",
    author = "Lester, Brian  and
      Al-Rfou, Rami  and
      Constant, Noah",
    editor = "Moens, Marie-Francine  and
      Huang, Xuanjing  and
      Specia, Lucia  and
      Yih, Scott Wen-tau",
    booktitle = "Proceedings of the 2021 Conference on Empirical Methods in Natural Language Processing",
    month = nov,
    year = "2021",
    address = "Online and Punta Cana, Dominican Republic",
    publisher = "Association for Computational Linguistics",
    url = "https://aclanthology.org/2021.emnlp-main.243/",
    doi = "10.18653/v1/2021.emnlp-main.243",
    pages = "3045--3059"
}

@inproceedings{li2021prefix,
  title        = {Prefix-Tuning: Optimizing Continuous Prompts for Generation},
  author       = {Li, Xiang Lisa and Liang, Percy},
  year         = 2021,
  booktitle    = {Proceedings of the 59th Annual Meeting of the Association for Computational Linguistics},
  pages        = {4582--4597},
}

@inproceedings{chen2021finqa,
    title = "{F}in{QA}: A Dataset of Numerical Reasoning over Financial Data",
    author = "Chen, Zhiyu  and
      Chen, Wenhu  and
      Smiley, Charese  and
      Shah, Sameena  and
      Borova, Iana  and
      Langdon, Dylan  and
      Moussa, Reema  and
      Beane, Matt  and
      Huang, Ting-Hao  and
      Routledge, Bryan  and
      Wang, William Yang",
    booktitle = "Proceedings of the 2021 Conference on Empirical Methods in Natural Language Processing",
    month = nov,
    year = "2021",
    address = "Online and Punta Cana, Dominican Republic",
    url = "https://aclanthology.org/2021.emnlp-main.300/",
    doi = "10.18653/v1/2021.emnlp-main.300",
    pages = "3697--3711",
}

@inproceedings{strich2026t2ragbench,
  title        = {{T2-RAGBench}: Text-and-Table Benchmark for Evaluating Retrieval-Augmented Generation},
  author       = {Strich, Jan and Isgorur, Enes Kutay and Trescher, Maximilian and Biemann, Chris and Semmann, Martin},
  year         = 2026,
  booktitle    = {Proceedings of the 19th Conference of the European Chapter of the Association for Computational Linguistics (Volume 1: Long Papers)},
  pages        = {165--191},
  url          = {https://aclanthology.org/2026.eacl-long.8/},
}

@article{adams2024longhealth,
  title        = {{LongHealth}: A Question Answering Benchmark with Long Clinical Documents},
  author       = {Adams, Lisa and Busch, Felix and Truhn, Daniel and Bressem, Keno K.},
  year         = 2024,
  journal      = {arXiv preprint arXiv:2401.14490},
}

@inproceedings{dasigi2021qasper,
  title        = {A Dataset of Information-Seeking Questions and Answers Anchored in Research Papers},
  author       = {Dasigi, Pradeep and Lo, Kyle and Beltagy, Iz and Cohan, Arman and Smith, Noah A. and Gardner, Matt},
  year         = 2021,
  booktitle    = {Proceedings of the 2021 Conference of the North American Chapter of the Association for Computational Linguistics},
  pages        = {4599--4610},
}

@article{pang2022quality,
  title        = {{QuALITY}: Question Answering with Long Input Texts, Yes!},
  author       = {Pang, Richard Yuanzhe and Parrish, Alicia and Joshi, Nitish and Nangia, Nikita and Phang, Jason and Chen, Angelica and Padmakumar, Vishakh and Ma, Johnny and Thompson, Jana and He, He and Bowman, Samuel R.},
  year         = 2022,
  journal      = {Proceedings of the 2022 Conference of the North American Chapter of the Association for Computational Linguistics},
  pages        = {5336--5358},
}

@inproceedings{castelli2020techqa,
    title = "The {T}ech{QA} Dataset",
    author = "Castelli, Vittorio  and
      Chakravarti, Rishav  and
      Dana, Saswati  and
      Ferritto, Anthony  and
      Florian, Radu  and
      Franz, Martin  and
      Garg, Dinesh  and
      Khandelwal, Dinesh  and
      McCarley, Scott  and
      McCawley, Michael  and
      Nasr, Mohamed  and
      Pan, Lin  and
      Pendus, Cezar  and
      Pitrelli, John  and
      Pujar, Saurabh  and
      Roukos, Salim  and
      Sakrajda, Andrzej  and
      Sil, Avi  and
      Uceda-Sosa, Rosario  and
      Ward, Todd  and
      Zhang, Rong",
    booktitle = "Proceedings of the 58th Annual Meeting of the Association for Computational Linguistics",
    month = jul,
    year = "2020",
    address = "Online",
    url = "https://aclanthology.org/2020.acl-main.117/",
    doi = "10.18653/v1/2020.acl-main.117",
    pages = "1269--1278",
}

@misc{eval-harness,
  title        = {The Language Model Evaluation Harness},
  author       = {Gao, Leo and Tow, Jonathan and Abbasi, Baber and Biderman, Stella and Black, Sid and DiPofi, Anthony and Foster, Charles and Golding, Laurence and Hsu, Jeffrey and Le Noac'h, Alain and Li, Haonan and McDonell, Kyle and Muennighoff, Niklas and Ociepa, Chris and Phang, Jason and Reynolds, Laria and Schoelkopf, Hailey and Skowron, Aviya and Sutawika, Lintang and Tang, Eric and Thite, Anish and Wang, Ben and Wang, Kevin and Zou, Andy},
  year         = 2024,
  doi          = {10.5281/zenodo.12608602},
  url          = {https://zenodo.org/records/12608602},
  version      = {v0.4.3},
}

@article{yang2025qwen3,
  title        = {{Qwen3 Technical Report}},
  author       = {Yang, An and Li, Anfeng and Yang, Baosong and Zhang, Beichen and Hui, Binyuan and Zheng, Bo and Yu, Bowen and Gao, Chang and Huang, Chengen and Lv, Chenxu and others},
  year         = 2025,
  journal      = {arXiv preprint arXiv:2505.09388},
  url          = {https://arxiv.org/abs/2505.09388},
}

@article{agarwal2025gpt,
  title        = {gpt-oss-120b \& gpt-oss-20b model card},
  author       = {Agarwal, Sandhini and Ahmad, Lama and Ai, Jason and Altman, Sam and Applebaum, Andy and Arbus, Edwin and Arora, Rahul K and Bai, Yu and Baker, Bowen and Bao, Haiming and others},
  year         = 2025,
  journal      = {arXiv preprint arXiv:2508.10925},
}

@article{guo2025deepseek,
  title        = {{DeepSeek-R1: Incentivizing Reasoning Capability in LLMs via Reinforcement Learning}},
  author       = {{DeepSeek-AI}},
  year         = 2025,
  journal      = {arXiv preprint arXiv:2501.12948},
  url          = {https://arxiv.org/abs/2501.12948},
}

@article{lewis2020retrieval,
  title        = {Retrieval-augmented generation for knowledge-intensive nlp tasks},
  author       = {Lewis, Patrick and Perez, Ethan and Piktus, Aleksandra and Petroni, Fabio and Karpukhin, Vladimir and Goyal, Naman and K{\"u}ttler, Heinrich and Lewis, Mike and Yih, Wen-tau and Rockt{\"a}schel, Tim and others},
  year         = 2020,
  journal      = {Advances in neural information processing systems},
  booktitle    = {Advances in Neural Information Processing Systems},
  volume       = 33,
  pages        = {9459--9474},
  url          = {https://proceedings.neurips.cc/paper_files/paper/2020/file/6b493230205f780e1bc26945df7481e5-Paper.pdf},
}

@inproceedings{lajewska-etal-2025-understanding,
  title        = "Understanding and Improving Information Preservation in Prompt Compression for {LLM}s",
  author       = "{\L}ajewska, Weronika  and
      Hardalov, Momchil  and
      Aina, Laura  and
      Anna John, Neha  and
      Su, Hang  and
      Marquez, Lluis",
  year         = 2025,
  booktitle    = "Findings of the Association for Computational Linguistics: EMNLP 2025",
  address      = "Suzhou, China",
  pages        = "17520--17541",
  doi          = "10.18653/v1/2025.findings-emnlp.949",
  isbn         = "979-8-89176-335-7",
  url          = "https://aclanthology.org/2025.findings-emnlp.949/",
}

@article{zweiger2026fast,
  title        = {Fast {KV} compaction via attention matching},
  author       = {Zweiger, Adam and Fu, Xinghong and Guo, Han and Kim, Yoon},
  year         = 2026,
  journal      = {arXiv preprint arXiv:2602.16284},
}

@article{gao2023retrieval,
  title        = {Retrieval-augmented generation for large language models: A survey},
  author       = {Gao, Yunfan and Xiong, Yun and Gao, Xinyu and Jia, Kangxiang and Pan, Jinliu and Bi, Yuxi and Dai, Yixin and Sun, Jiawei and Wang, Haofen and Wang, Haofen and others},
  year         = 2023,
  journal      = {arXiv preprint arXiv:2312.10997},
  volume       = 2,
  number       = 1,
  pages        = 32,
}

@inproceedings{guu2020retrieval,
  title        = {Retrieval augmented language model pre-training},
  author       = {Guu, Kelvin and Lee, Kenton and Tung, Zora and Pasupat, Panupong and Chang, Mingwei},
  year         = 2020,
  booktitle    = {International conference on machine learning},
  pages        = {3929--3938},
}

@inproceedings{yadav2023tiesmerging,
  title={{TIES}-Merging: Resolving Interference When Merging Models},
  author={Yadav, Prateek and Tam, Derek and Choshen, Leshem and Raffel, Colin and Bansal, Mohit},
  booktitle={Advances in Neural Information Processing Systems},
  year={2023}
}

@article{yang2026kvlink,
  title={Kvlink: Accelerating large language models via efficient kv cache reuse},
  author={Yang, Jingbo and Hou, Bairu and Wei, Wei and Bao, Yujia and Chang, Shiyu},
  journal={Advances in Neural Information Processing Systems},
  volume={38},
  pages={133797--133824},
  year={2026}
}

@article{chen2026kv,
  title={KV Packet: Recomputation-Free Context-Independent KV Caching for LLMs},
  author={Chen, Chuangtao and Zhang, Grace Li and Yin, Xunzhao and Zhuo, Cheng and Li, Bing and Schlichtmann, Ulf},
  journal={arXiv preprint arXiv:2604.13226},
  year={2026}
}

@inproceedings{asai2024self,
  title={Self-rag: Learning to retrieve, generate, and critique through self-reflection},
  author={Asai, Akari and Wu, Zeqiu and Wang, Yizhong and Sil, Avi and Hajishirzi, Hannaneh},
  booktitle={International conference on learning representations},
  volume={2024},
  pages={9112--9141},
  year={2024}
}

@article{du2026c,
  title={{C\textsuperscript{2}KV}: Compressed and Composable KV Cache Reuse for Efficient LLM Inference},
  author={Du, Chuheng and Chen, Junyi and Tang, Hanlin and Liu, Kan and Lan, Tao and Qu, Lin and Niu, Chaoyue and Liu, Shengzhong and Chen, Guihai and Wu, Fan},
  journal={arXiv preprint arXiv:2607.17715},
  year={2026}
}

@inproceedings{yu2024dare,
  title={Language Models are Super Mario: Absorbing Abilities from Homologous Models as a Free Lunch},
  author={Yu, Le and Yu, Bowen and Yu, Haiyang and Huang, Fei and Li, Yongbin},
  booktitle={International Conference on Machine Learning},
  year={2024}
}

@article{cobbe2021gsm8k,
  title={Training Verifiers to Solve Math Word Problems},
  author={Cobbe, Karl and Kosaraju, Vineet and Bavarian, Mohammad and Chen, Mark and Jun, Heewoo and Kaiser, Lukasz and Plappert, Matthias and Tworek, Jerry and Hilton, Jacob and Nakano, Reiichiro and Hesse, Christopher and Schulman, John},
  journal={arXiv preprint arXiv:2110.14168},
  year={2021}
}

@article{chen2021humaneval,
  title={Evaluating Large Language Models Trained on Code},
  author={Chen, Mark and Tworek, Jerry and Jun, Heewoo and Yuan, Qiming and Pinto, Henrique Ponde de Oliveira and Kaplan, Jared and Edwards, Harri and Burda, Yuri and Joseph, Nicholas and Brockman, Greg and others},
  journal={arXiv preprint arXiv:2107.03374},
  year={2021}
}

@article{zhou2023ifeval,
  title={Instruction-Following Evaluation for Large Language Models},
  author={Zhou, Jeffrey and Lu, Tianjian and Mishra, Swaroop and Brahma, Siddhartha and Basu, Sujoy and Luan, Yi and Zhou, Denny and Hou, Le},
  journal={arXiv preprint arXiv:2311.07911},
  year={2023}
}

@inproceedings{hendrycks2021mmlu,
  title={Measuring Massive Multitask Language Understanding},
  author={Hendrycks, Dan and Burns, Collin and Basart, Steven and Zou, Andy and Mazeika, Mantas and Song, Dawn and Steinhardt, Jacob},
  booktitle={International Conference on Learning Representations (ICLR)},
  year={2021}
}

@inproceedings{ovadia2024finetuning,
  title={Fine-Tuning or Retrieval? Comparing Knowledge Injection in {LLM}s},
  author={Ovadia, Oded and Brief, Menachem and Mishaeli, Moshik and Elisha, Oren},
  booktitle={Proceedings of the 2024 Conference on Empirical Methods in Natural Language Processing (EMNLP)},
  year={2024}
}

@article{tang2025understanding,
  title={Understanding Parametric Knowledge Injection in Retrieval-Augmented Generation},
  author={Tang, Yichen and Su, Weihang and Ai, Qingyao and Liu, Yiqun},
  journal={arXiv preprint arXiv:2510.12668},
  year={2025}
}

@inproceedings{NEURIPS2020_1457c0d6,
 author = {Brown, Tom and Mann, Benjamin and Ryder, Nick and Subbiah, Melanie and Kaplan, Jared D and Dhariwal, Prafulla and Neelakantan, Arvind and Shyam, Pranav and Sastry, Girish and Askell, Amanda and Agarwal, Sandhini and Herbert-Voss, Ariel and Krueger, Gretchen and Henighan, Tom and Child, Rewon and Ramesh, Aditya and Ziegler, Daniel and Wu, Jeffrey and Winter, Clemens and Hesse, Chris and Chen, Mark and Sigler, Eric and Litwin, Mateusz and Gray, Scott and Chess, Benjamin and Clark, Jack and Berner, Christopher and McCandlish, Sam and Radford, Alec and Sutskever, Ilya and Amodei, Dario},
 booktitle = {Advances in Neural Information Processing Systems},
 NOeditor = {H. Larochelle and M. Ranzato and R. Hadsell and M.F. Balcan and H. Lin},
 pages = {1877--1901},
 NOpublisher = {Curran Associates, Inc.},
 title = {Language Models are Few-Shot Learners},
 url = {https://proceedings.neurips.cc/paper_files/paper/2020/file/1457c0d6bfcb4967418bfb8ac142f64a-Paper.pdf},
 volume = {33},
 year = {2020}
}
\clearpage

\appendix

\section{Hyperparameters}
\label{sec:hyperparameters}
 We follow the hyperparameters in \citep{hardalov2026cartridges} for Cartridges. We train for 20 epochs on QASPER, FinQA, and TechQA, and 5 epochs on LongHealth and QuALITY. The longer-document datasets require more epochs to converge. All parametric methods use the AdamW optimizer with gradient clipping at $1.0$, and an effective batch size of $64$ (via gradient accumulation). We found that using learning-rate schedule or warmup did not result in any performance boost. LoRA uses rank $r{=}64$ and $\alpha{=}128$, applied to the feed-forward projections (gate, up, down) of every layer, at learning rate $1\times10^{-4}$. The MLP adapter has a bottleneck dimension of $d{=}512$ inserted as a residual on every layer's output, also at $1\times10^{-4}$. The two largest variants ($d\in\{1536,3072\}$) instead use $2\times10^{-5}$, as $1\times10^{-4}$ was training-unstable at that size. We set the learning rate to $2\times10^{-5}$ for full fine-tuning.

\section{LoRA Merging Approaches}
\label{sec:merging}
In the multi-document setting we compose per-document LoRA adapters by merging the top-$k$ retrieved adapters per question. The main results use a linear average (\emph{mean}). Here we compare it against three alternatives from the model-merging literature: \emph{cat}, which concatenates the low-rank factors; \emph{TIES}~\citep{yadav2023tiesmerging}, which trims each adapter to its highest-magnitude entries (density $0.5$), resolves sign conflicts, and averages the agreeing weights; and \emph{DARE}~\citep{yu2024dare}, which randomly drops and rescales weights before a linear merge. We compare the merging methods with top-$5$ retrieved documents in Table~\ref{tab:merging}. The four merge strategies are within roughly one point of one another on every dataset. The failure is therefore in composing independently trained adapters at all, not in the choice of merge operator.

\section{Gemma-3-12B Results}
\label{sec:gemma}
To check that our findings are not specific to Qwen3-8B, we replicate the single-document setting on Gemma-3-12B~\cite{gemma3_report} (Table~\ref{tab:gemma-single-doc}). Similar to Qwen3-8B, \emph{No context} is far below all knowledge injection methods, and Compaction at $2\times$ nearly matches the in-context oracle ($65.2$ vs.\ $68.9$) while the parametric LoRA trails behind ($50.8$). Unlike on Qwen3-8B, the trained Cartridges underperform here ($37.3$ on average): on Gemma-3-12B they stay below the training-free Compaction on every dataset, which we attribute to their catastrophic forgetting as presented in Table~\ref{tab:gemma-forgetting}.

\begin{table}[t]
\centering
\small
\begin{tabular}{lcccc}
\toprule
\textbf{Merge} & \textbf{LongH.} & \textbf{QuAL.} & \textbf{FinQA} & \textbf{TechQA} \\
\midrule
Mean        & 51.2 {\scriptsize $\pm$ 1.9} & 56.0 {\scriptsize $\pm$ 0.7} & 5.1 {\scriptsize $\pm$ 0.1} & 21.7 {\scriptsize $\pm$ 0.4} \\
Cat         & 51.0 {\scriptsize $\pm$ 2.2} & 56.0 {\scriptsize $\pm$ 0.7} & 5.0 {\scriptsize $\pm$ 0.1} & 22.1 {\scriptsize $\pm$ 1.0} \\
TIES        & 51.2 {\scriptsize $\pm$ 2.3} & 56.0 {\scriptsize $\pm$ 0.6} & 5.1 {\scriptsize $\pm$ 0.1} & 21.9 {\scriptsize $\pm$ 0.9} \\
DARE        & 50.9 {\scriptsize $\pm$ 2.0} & 56.0 {\scriptsize $\pm$ 0.7} & 5.0 {\scriptsize $\pm$ 0.1} & 22.5 {\scriptsize $\pm$ 0.8} \\
\bottomrule
\end{tabular}
\caption{\textbf{LoRA merging approaches} (Qwen3-8B, rank $64$). We merge the adapters of the five retrieved documents per question. Scores are the mean $\pm$ standard deviation over 3 seeds.}
\label{tab:merging}
\end{table}

\begin{table}[t]
\centering
\resizebox{\columnwidth}{!}{%
\small
\setlength{\tabcolsep}{3pt}
\begin{tabular}{lcccccc}
\toprule
\textbf{Method} & \textbf{LongHealth} & \textbf{QuALITY} & \textbf{QASPER} & \textbf{FinQA} & \textbf{TechQA} & \textbf{Avg.} \\
 & (Acc.) & (Acc.) & (F1) & (EM) & (Judge) & \\
\midrule
No context         & 40.6 {\scriptsize $\pm$ 1.4} & 28.0 {\scriptsize $\pm$ 0.7} & 19.3 {\scriptsize $\pm$ 0.3} & 5.2 {\scriptsize $\pm$ 0.0} & 14.8 {\scriptsize $\pm$ 1.2} & 21.6 \\
ICL (oracle)       & \textbf{78.5} {\scriptsize $\pm$ 0.4} & \textbf{77.4} {\scriptsize $\pm$ 0.2} & \textbf{51.3} {\scriptsize $\pm$ 0.3} & \textbf{61.9} {\scriptsize $\pm$ 0.0} & \textbf{75.4} {\scriptsize $\pm$ 1.9} & \textbf{68.9} \\
\cmidrule(l){1-7}
Cartridge   & 49.8 {\scriptsize $\pm$ 2.1} & 43.0 {\scriptsize $\pm$ 1.4} & 23.4 {\scriptsize $\pm$ 1.2} & 9.5 {\scriptsize $\pm$ 1.7} & 60.6 {\scriptsize $\pm$ 0.3} & 37.3 \\
Compaction  & \textbf{81.4} {\scriptsize $\pm$ 0.3} & 74.4 {\scriptsize $\pm$ 0.5} & 48.0 {\scriptsize $\pm$ 0.2} & 61.2 {\scriptsize $\pm$ 0.9} & 61.0 {\scriptsize $\pm$ 2.6} & 65.2 \\
LoRA               & 56.5 {\scriptsize $\pm$ 1.2} & 62.4 {\scriptsize $\pm$ 0.5} & 34.8 {\scriptsize $\pm$ 1.0} & 34.7 {\scriptsize $\pm$ 0.2} & 65.6 {\scriptsize $\pm$ 1.0} & 50.8 \\
\bottomrule
\end{tabular}%
}
\caption{\textbf{Single-document results on Gemma-3-12B.} Same protocol as Table~\ref{tab:single-doc} (Compaction $2\times$, LoRA rank $64$. Scores averaged over 3 runs $\pm$ standard deviation where available).}
\label{tab:gemma-single-doc}
\end{table}

\begin{table}[t]
\centering
\resizebox{\columnwidth}{!}{%
\small
\begin{tabular}{lcccc}
\toprule
\textbf{Method} & \textbf{GSM8K} & \textbf{HumanEval} & \textbf{IFEval} & \textbf{MMLU} \\
\midrule
Base model        & \textbf{88.7} & \textbf{83.5} & \textbf{78.4} & \textbf{72.6} \\
\cmidrule(l){1-5}
Cartridge & 36.1 {\scriptsize $\pm$ 23.9} & 54.4 {\scriptsize $\pm$ 14.6} & 39.2 {\scriptsize $\pm$ 15.7} & 56.9 {\scriptsize $\pm$ 8.2} \\
\bottomrule
\end{tabular}%
}
\caption{\textbf{Forgetting on control benchmarks (Gemma-3-12B).} Base-model accuracy followed by the $2\times$ trained Cartridges.}
\label{tab:gemma-forgetting}
\end{table}

\section{Distillation vs.\ Next-token Prediction}
\label{sec:distill-ntp}
Our training objective distills the teacher's next-token distribution rather than training on the teacher's sampled tokens with the standard next-token-prediction cross-entropy. Table~\ref{tab:distill-ntp} compares the two objectives for LoRA in the single-document setting, holding everything else fixed. Distillation outperforms next-token-prediction on every dataset and is close to $4$ points higher on average, with the largest gains on the reasoning-heavy datasets. This corroborates the objective choice of \citet{eyuboglu2025cartridges} and motivates its use throughout our experiments.

\begin{table}[t]
\centering
\small
\begin{tabular}{lcc}
\toprule
\textbf{Dataset} & \textbf{Distillation} & \textbf{Next-token prediction} \\
\midrule
LongHealth & \textbf{75.2} {\scriptsize $\pm$ 0.9} & 72.6 {\scriptsize $\pm$ 1.6} \\
QuALITY    & \textbf{73.7} {\scriptsize $\pm$ 1.2} & 70.7 {\scriptsize $\pm$ 0.7} \\
QASPER     & \textbf{50.3} {\scriptsize $\pm$ 0.6} & 46.9 {\scriptsize $\pm$ 1.0} \\
FinQA      & \textbf{49.0} {\scriptsize $\pm$ 0.9} & 41.3 {\scriptsize $\pm$ 0.6} \\
TechQA     & \textbf{74.8} {\scriptsize $\pm$ 3.3} & 73.2 {\scriptsize $\pm$ 1.2} \\
\midrule
Avg.       & \textbf{64.6} & 60.9 \\
\bottomrule
\end{tabular}
\caption{\textbf{Distillation vs.\ next-token prediction} for per-document LoRA (Qwen3-8B, rank $64$), single-document setting. Scores are the mean $\pm$ standard deviation over 3 seeds.}
\label{tab:distill-ntp}
\end{table}

\section{Evaluation Prompts}
\label{sec:eval_prompts}

We use task-specific system prompts during evaluation. All prompts instruct the model to wrap its final answer in \texttt{<answer>} tags. In the Cartridge setting, the document context is encoded in the trained KV cache prefix and is not included in the text prompt, so the model receives only the system prompt and the user question. In the Oracle (Text) baseline, the full document text is prepended to the user message. Table~\ref{tab:eval-prompts} lists the exact prompts used.

For FinQA, we use the evaluation prompt recommended in the T$^2$-RAGBench paper~\cite{strich2026t2ragbench}.\footnote{\url{https://github.com/uhh-hcds/g4kmu-paper/tree/main/src/g4k/prompts/sys_prompts}} For QASPER\footnote{\url{https://github.com/EleutherAI/lm-evaluation-harness/tree/main/lm_eval/tasks/qasper}} and QuALITY, we adopt the prompts from previous work~\cite{eval-harness,eyuboglu2025cartridges,zweiger2026fast}. For LongHealth, we use the prompt format from the original benchmark\footnote{\url{https://github.com/HazyResearch/cartridges/blob/main/examples/benchmarks/longhealth/baseline_longhealth.py}}. For TechQA, since no standard evaluation prompt exists, we conducted a prompt search over several variants, including a minimal instruction (``Answer concisely''), a domain-specific expert framing, a chain-of-thought variant, and a structured extraction format, and selected the prompt that yielded the highest Oracle (Text) accuracy on the development set.

\begin{table*}[t]
\centering
\footnotesize
\begin{tabular}{@{}p{0.15\textwidth}p{0.82\textwidth}@{}}
\toprule
\textbf{Benchmark} & \textbf{System Prompt + User Message Format} \\
\midrule
T$^2$-RAGBench / FinQA &
\texttt{System: You are an expert in answering financial questions by constructing mathematical formulas based on a simple syntax.} \\
& \texttt{- Task: Provide a FORMULA ANSWER to the question based on the given context.} \\
& \texttt{Guidelines:} \\
& \texttt{1. Answer Type: A formula is either a number or one of: add(f1, f2), subtract(f1, f2), multiply(f1, f2), divide(f1, f2), exp(f1, f2), greater(f1, f2)} \\
& \texttt{2. Reasoning: Carefully analyze the context. Pay special attention to the table.} \\
& \texttt{3. Final Answer: "<answer> FORMULA </answer>"} \\
& \\
& \texttt{User: \{context\}} \\
& \texttt{Question: \{question\}} \\
\midrule
LongHealth &
\texttt{System: Please reference the patient medical records to answer the user's questions. Choose the single best option and provide your answer exactly as it appears in the options.} \\
& \texttt{Wrap your answer in: <answer> The correct option text here </answer>} \\
& \\
& \texttt{User: \{question\}} \\
& \texttt{A. \{option\_a\}~~B. \{option\_b\}~~C. \{option\_c\}~~D. \{option\_d\}~~E. \{option\_e\}} \\
\midrule
QASPER &
\texttt{System: You are a research assistant answering questions about a scientific paper. Answer as briefly as possible. Give only the answer, no explanation. If the question cannot be answered from the paper, say "Unanswerable". For yes/no questions, answer "Yes" or "No". Wrap your answer in: <answer> ... </answer>} \\
& \\
& \texttt{User: \{question\}} \\
\midrule
QuALITY &
\texttt{System: You are a careful reader answering multiple-choice questions about a long article. Read the article and choose the single best answer option. Provide your answer as the letter (A, B, C, or D) wrapped in answer tags. Example: <answer> B </answer>} \\
& \\
& \texttt{User: \{question\}} \\
& \texttt{A. \{option\_a\}~~B. \{option\_b\}~~C. \{option\_c\}~~D. \{option\_d\}} \\
& \texttt{Answer with the letter of the correct option (A, B, C, or D).} \\
\midrule
TechQA &
\texttt{System: You are an expert technical support assistant specializing in IT infrastructure, software products, and enterprise systems. Answer the technical question to the best of your ability. Be concise and factual. Wrap your answer in: <answer> YOUR\_ANSWER </answer>} \\
& \\
& \texttt{User: \{question\}} \\
\bottomrule
\end{tabular}
\caption{Evaluation prompts used for each benchmark. Variables in braces are replaced with benchmark data at evaluation time.}
\label{tab:eval-prompts}
\end{table*}

\section{LLM-as-a-Judge evaluation}
\label{sec:judge}
A separate judge model (DeepSeek-Distilled-Qwen-32B) evaluates whether the TechQA generated answer is factually equivalent to the reference answer. The judge outputs a JSON object with a justification and a binary grade. A score of 1 is assigned when the judge returns \texttt{"grade": "correct"}, and 0 otherwise. Table~\ref{tab:llmaaj-prompt} shows the exact judge prompt.

\begin{table*}[t]
\centering
\footnotesize
\begin{tabular}{@{}p{0.12\textwidth}p{0.85\textwidth}@{}}
\toprule
\textbf{Role} & \textbf{Prompt} \\
\midrule
System &
\texttt{You are to act as an impartial judge, evaluating whether an answer to a question matches a provided reference answer. Your task is to determine if the given answer is correct based on its factual equivalence to the reference answer, ignoring differences in punctuation and phrasing.} \\
& \\
& \texttt{When evaluating the answer, consider the following criteria:} \\
& \texttt{1. Factual equivalence: Does the answer convey the same information as the reference answer?} \\
& \texttt{2. Completeness: Does the answer address all parts of the question that the reference answer addresses?} \\
& \texttt{3. Accuracy: Is the information in the answer consistent with the reference answer?} \\
& \texttt{4. Additional information: If the answer contains more information than the reference answer, does it remain consistent and not contradict itself?} \\
& \\
& \texttt{Your response should be structured as follows:} \\
& \texttt{1. A justification for your decision, explaining your reasoning based on the evaluation criteria.} \\
& \texttt{2. A grade of either "correct" or "incorrect".} \\
& \\
& \texttt{Important note on deflections and invalid questions:} \\
& \texttt{- If the answer is a deflection or does not attempt to answer the question, grade it as "incorrect" unless the reference answer is also a deflection.} \\
& \texttt{- If both the answer and the reference answer indicate that the question is invalid or cannot be answered, grade it as "correct".} \\
& \texttt{- If the answer is a placeholder like \{\{YOUR\_ANSWER\}\}, any generic phrase that does NOT address the question, or an empty answer, then count it as "incorrect".} \\
& \\
& \texttt{Your response should be in json format as follows:} \\
& \texttt{\{"justification": "...", "grade": "correct" or "incorrect"\}} \\
\midrule
User &
\texttt{Here is the question:} \\
& \texttt{\{query\}} \\
& \\
& \texttt{Here is the answer to be judged:} \\
& \texttt{\{answer\}} \\
& \\
& \texttt{Here is the reference answer:} \\
& \texttt{\{expected\_answer\}} \\
\bottomrule
\end{tabular}
\caption{LLM-as-a-Judge prompt used for scoring free-form answers on TechQA. The judge model (DeepSeek-Distilled-Qwen-32B) receives the system prompt defining evaluation criteria and a user message containing the question, generated answer, and reference answer.}
\label{tab:llmaaj-prompt}
\end{table*}

\section{Self-Study Data Synthesis Details}
\label{sec:synthesis_details}

We generate the self-study data following \citet{eyuboglu2025cartridges,hardalov2026cartridges}.

\paragraph{Question generation model.}
We use GPT-OSS 120B~\cite{agarwal2025gpt} as the question generator ($M_Q$).
This model receives the full document context in its system prompt and generates batches of 20 diverse questions per call.
The answer generator ($M_A$) is the target model itself (Qwen3-8B), ensuring that the distillation signal reflects the student's own output distribution.

\paragraph{Seed prompt types.}
Each synthesis call randomly selects a seed prompt type from five categories: \emph{structuring} (requests to organize information into JSON, YAML, or other formats), \emph{summarization} (requests to summarize specific sections), \emph{question} (factual recall and reasoning questions), \emph{use\_case} (practical downstream application tasks), and \emph{creative} (open-ended discussion prompts).
This diversity ensures the cartridge is trained on varied interaction patterns rather than only factoid QA.

\paragraph{Multi-question prompt format.}
In batched mode, the question generator receives the following instruction (shown for the \emph{question} type):

\begin{quote}
\small
\texttt{Generate \{n\} diverse questions that test knowledge of the information in the corpus above. Each question should cover a different fact, detail, or aspect of the corpus. Vary the style: mix factual recall, comparison, reasoning, and detail-oriented questions. Include specific details (ids, names, titles, dates, numerical values, etc.) in each question so it is clear what you are asking about.}

\texttt{Output ONLY a JSON array of strings, e.g.,~["question 1", "question 2", ...]. No other text, no markdown fences, no explanation.}
\end{quote}

The structured JSON output format enables reliable parsing; responses that fail to parse are discarded (typically ${<}5\%$ of calls).

\paragraph{Proportional-to-length sampling.}
We replace uniform random chunk sampling with a \emph{proportional-to-length} strategy ensuring balanced document coverage.
Each document $d_i$ in the collection is assigned a sampling weight $w_i = \frac{|d_i|}{\min_j |d_j|}$, so that longer documents, which are assumed to contain more facts, are sampled proportionally more often during synthesis.

\paragraph{Sampling rounds and temperatures.}
For each dataset, we run 4 independent sampling rounds, one per seed prompt type (question, structuring, summarization, use\_case), each generating 10{,}000 samples for a total of 40{,}000 training examples per dataset.\footnote{We exclude the \emph{creative} seed type from training data as it produces open-ended prompts less suited for distillation.}
$M_Q$ (GPT-OSS 120B, the question generator) uses temperature $0.6$ with top-$p{=}0.95$, top-$k{=}20$, and a maximum of 4{,}096 completion tokens per call.
$M_A$ (Qwen3-8B, the teacher/answer generator) uses temperature $0.0$ (greedy decoding) with a maximum of 2{,}048 completion tokens, ensuring deterministic distillation targets.
Both models operate with thinking mode enabled.

\paragraph{Sampling configuration.}
Within each round, we generate 10{,}000 synthesis samples with a batch size of 4 contexts and up to 128 parallel API calls.
Documents are sampled with a fixed random seed for reproducibility.
For multi-note documents (e.g., LongHealth patient records), we sample one note per prompt to ensure fine-grained coverage of individual clinical notes.

\section{RAG Baseline: Indexing and Retrieval Details}
\label{sec:rag_details}

We describe the full RAG pipeline below.

\paragraph{Document indexing.}
Each dataset's documents are indexed with chunk size $C = 512$ tokens.
Documents are chunked using a fixed-size strategy with a 10\% token overlap between adjacent chunks.
Chunks are embedded using Amazon Titan Embed Text v2 (\texttt{amazon.titan-embed-text-v2:0}) via Amazon Bedrock\footnote{\url{https://aws.amazon.com/bedrock}} with 1024-dimensional embeddings and retrieved via cosine similarity.

\paragraph{Retrieval.}
At evaluation time, each question is issued as a retrieval query against the Knowledge Base.
We retrieve the top-$K$ chunks for $K \in \{1, 3, 5, 10\}$.
To avoid redundant API calls, we retrieve $K = 10$ once per chunk size and slice for smaller $K$ values.
Retrieved chunks are concatenated in score order and prepended to the question as the context for the reader model.
Table~\ref{tab:unique_docs} reports how many unique source documents the top-$k$ retrieved chunks cover.

\begin{table}[t]
\centering
\small
\begin{tabular}{lcccc}
\toprule
\textbf{Dataset} & $k{=}1$ & $k{=}3$ & $k{=}5$ & $k{=}10$ \\
\midrule
LongHealth & 1.0 & 1.8 & 2.5 & 4.4 \\
QuALITY    & 1.0 & 1.4 & 2.2 & 4.8 \\
FinQA      & 1.0 & 3.0 & 5.0 & 8.7 \\
TechQA     & 1.0 & 3.0 & 4.9 & 8.1 \\
\bottomrule
\end{tabular}
\caption{Average number of unique source documents covered by the top-$k$ retrieved chunks (chunk size 1024). Datasets with shorter documents (FinQA, TechQA) have higher chunk diversity, while longer-document datasets (LongHealth, QuALITY) exhibit within-document clustering.}
\label{tab:unique_docs}
\end{table}

\paragraph{Effective token counts.}
Table~\ref{tab:rag_tokens} reports the average number of context tokens consumed per query for Text RAG at chunk size 1024 across different $k$ values. Due to the 10\% overlap between adjacent chunks and shorter final chunks in some documents, the effective token count is slightly below the nominal $k \times 1024$.

\begin{table}[t]
\centering
\small
\begin{tabular}{lcccc}
\toprule
\textbf{Dataset} & $k{=}1$ & $k{=}3$ & $k{=}5$ & $k{=}10$ \\
\midrule
LongHealth & 986 & 2{,}960 & 4{,}944 & 9{,}860 \\
QuALITY    & 941 & 2{,}856 & 4{,}733 & 9{,}412 \\
FinQA      & ${\sim}$960 & ${\sim}$2{,}880 & ${\sim}$4{,}800 & ${\sim}$9{,}600 \\
TechQA     & ${\sim}$960 & ${\sim}$2{,}880 & ${\sim}$4{,}800 & ${\sim}$9{,}600 \\
\bottomrule
\end{tabular}
\caption{Average context tokens per query for Text RAG (chunk size 1024).}
\label{tab:rag_tokens}
\end{table}

\paragraph{Retrieval Quality: Recall and MRR.}
\label{sec:rag_recall}
We use the same document-level retrieval as \citet{hardalov2026cartridges}. Across all four datasets (LongHealth, QuALITY, FinQA, and TechQA), the dense retriever achieves consistently high retrieval quality, with Recall@10 reaching approximately 95\% or higher across chunk sizes, indicating that the relevant documents are almost always retrieved. This suggests that remaining performance differences are primarily due to information loss from chunking rather than retrieval failures. For the complete retrieval methodology, detailed Recall@K and MRR@K results, and further analysis, can be found in Appendix G \cite{hardalov2026cartridges}.

\section{Artifact Licenses and Terms of Use}
\label{appx:licenses}

\paragraph{Licenses and terms for use/distribution of artifacts.}

The five knowledge-intensive benchmarks used in this work are publicly available under the following licenses:

\begin{itemize}[nosep]
    \item \textbf{LongHealth}~\cite{adams2024longhealth}: Released under the Apache-2.0 License.\footnote{\url{https://github.com/kbressem/LongHealth/blob/main/LICENSE}} The dataset consists of entirely fictional patient records created by the authors; no real patient data is included.
    \item \textbf{QASPER}~\cite{dasigi2021qasper}: Released under the CC BY 4.0 License.\footnote{\url{https://huggingface.co/datasets/allenai/qasper}} The dataset consists of questions and answers over NLP research papers; the dataset itself is distributed under CC BY 4.0.
    \item \textbf{QuALITY}~\cite{pang2022quality}: Released under the CC BY 4.0 License.\footnote{\url{https://nyu-mll.github.io/quality/}} Source texts are drawn from Project Gutenberg (public domain) and other permissively licensed collections, including nonfiction and fiction sources such as Slate articles from the Open American National Corpus, The Long+Short, Freesouls, and Open Access books. These texts are published works and do not contain personal data, though some may include mature themes typical of literary and journalistic content.
    \item \textbf{FinQA}~\cite{chen2021finqa}: Released under the MIT License.\footnote{\url{https://github.com/czyssrs/FinQA/blob/main/LICENSE}} The underlying financial reports are sourced from corporate filings made available via the SEC EDGAR system, including earnings reports and annual/quarterly reports (e.g., 10-K and 10-Q documents). These are publicly accessible financial disclosures prepared by reporting companies. T$^2$-RAGBench~\cite{strich2026t2ragbench} is also released under the MIT License.
    \item \textbf{TechQA}~\cite{castelli2020techqa}: The NVIDIA TechQA-RAG-Eval variant used in this work is released under the Apache-2.0 License and is explicitly cleared for commercial and non-commercial use.\footnote{\url{https://huggingface.co/datasets/nvidia/TechQA-RAG-Eval}} It is derived from the original IBM TechQA dataset, whose code repository is also Apache-2.0.\footnote{\url{https://github.com/IBM/techqa/blob/master/LICENSE.md}}
\end{itemize}

We additionally use four control benchmarks to measure catastrophic forgetting, all publicly available:

\begin{itemize}[nosep]
    \item \textbf{GSM8K}~\cite{cobbe2021gsm8k}: grade-school math word problems, released under the MIT License.
    \item \textbf{HumanEval}~\cite{chen2021humaneval}: Python code-generation problems, released under the MIT License.
    \item \textbf{IFEval}~\cite{zhou2023ifeval}: instruction-following prompts, released under the Apache-2.0 License.
    \item \textbf{MMLU}~\cite{hendrycks2021mmlu}: multiple-choice questions across 57 subjects, released under the MIT License.
\end{itemize}

The models used in our experiments, \textbf{Qwen3-8B}~\cite{yang2025qwen3} and \textbf{Gemma-3-12B}~\cite{gemma3_report}, are released under the Apache 2.0 License and the Gemma Terms of Use respectively, both of which permit research use.

We do not release new datasets in this work. The trained cartridge artifacts (KV cache parameters) are derivatives of the above datasets and the model weights; any release of such artifacts would be subject to the intersection of the applicable licenses above.

\paragraph{Offensive content and personal data.}

\begin{itemize}[nosep]
    \item \textbf{LongHealth}: The dataset consists of entirely fictional patient records with no real individuals. No anonymization was required. We verified that no real names, addresses, or identifying information appear in the data.
    \item \textbf{QASPER}: Source texts are NLP research papers. No personal data or offensive content is expected; the domain is scientific writing.
    \item \textbf{QuALITY}: Source texts are fiction and non-fiction narratives, these are published literary works; no personal data is present.
    \item \textbf{FinQA / T$^2$-RAGBench}: Source texts are corporate earnings reports from SEC EDGAR. These are formal financial documents; no personal data or offensive content is present.
    \item \textbf{TechQA}: Source texts are IBM Technote IT support documents. These are technical documentation; no personal data or offensive content is expected.
    \item \textbf{Control benchmarks (GSM8K, HumanEval, IFEval, MMLU)}: These are widely used, publicly released evaluation suites of math word problems, programming problems, instruction-following prompts, and academic multiple-choice questions. They contain no personal data, and no offensive content is expected.
\end{itemize}

No additional anonymization steps were taken beyond those applied by the original dataset creators, as none of the datasets contain personal data about private individuals.

\section{AI use disclosure} We used AI to assist with code writing and manuscript typesetting. 

\end{document}